\documentclass[runningheads]{llncs}
\usepackage[utf8]{inputenc}
\usepackage{graphicx}
\usepackage{url}
\usepackage{afterpage}

\newcommand{\system}[0]{{\textit{EventKG+BT}}}

\begin{document}
\title{EventKG+BT: Generation of Interactive Biography Timelines from a Knowledge Graph}

\titlerunning{EventKG+BT}

\author{Simon Gottschalk\orcidID{0000-0003-2576-4640} \and \\
Elena Demidova\orcidID{0000-0002-5134-9072}}
\authorrunning{S. Gottschalk and E. Demidova}

\institute{L3S Research Center, Leibniz Universität Hannover, Germany\\
\email{\{gottschalk,demidova\}@L3S.de}}
\maketitle              
\begin{abstract}
Research on notable accomplishments and important events in the life of people of public interest usually requires close reading of long encyclopedic or biographical sources, which is a tedious and time-consuming task. 
Whereas semantic reference sources, such as the EventKG knowledge graph,
provide structured representations of relevant facts, they often include hundreds of events and temporal relations for particular entities.
In this paper, we present EventKG+BT -- a timeline generation system that creates concise and interactive spatio-temporal representations of biographies from a knowledge graph using distant supervision. 

\end{abstract}

\section{Introduction}
\label{sec:introduction}

Wikipedia, with more than one million articles dedicated to famous people, as well as other encyclopedic or biographical corpora on the Web, are rich sources of biographical information. 
These sources can help to answer questions like \textit{``What were the notable accomplishments in the life of George Washington?''}, and to learn about the life of people of public interest. 
Researchers who analyse event-centric cross-lingual information (in particular, computer scientists, information designers, and sociologists) prefer to approach such questions by exploiting concise representations, rather than by close reading of long articles \cite{Gottschalk:2018TPDL}. 

In this paper, we introduce \system{}\footnote{\url{http://eventkg-biographies.l3s.uni-hannover.de}} -- a system that enables exploration of personal biographies based on concise biography timelines.
In \cite{gottschalk2019eventkg}, we have shown how to automatically extract biography timelines from EventKG, an event-centric and temporal knowledge graph \cite{GottschalkD18}, using a distant supervision approach. 
In this approach, we trained an SVM classifier to predict the relevance of the potential timeline entries to a biography. 
We obtained training data through the mapping of facts extracted from biographical articles
to the temporal relations in the EventKG knowledge graph. 

We demonstrate the \system{} system that implements the distant supervision approach to biography timeline generation presented in \cite{gottschalk2019eventkg} and presents the results of this approach on an interactive biography timeline. 
We illustrate how biography timelines generated by \system{} 
can help to obtain a concise overview of a biography, alleviating the burden of time-consuming reading of long biographical or encyclopedic articles. 

\section{Biography Timelines}
\label{sec:interface}

We assume a use case where the user task is to gain insights into the life of a person of interest, e.g., to get the first impression and a rough understanding of that person's role in history, the notable accomplishments, and to obtain a starting point for further in-depth research. To this extent, \system{} shows a \textit{biography timeline} to the user as the core of the visualisation. As defined in \cite{gottschalk2019eventkg}, a biography timeline is a chronologically ordered list of temporal relations involving the person of interest:

\begin{definition}
A biography timeline $TL(p, bio) = (r_1,\dots,r_n)$ of a person of interest $p$ is a chronologically ordered list of timeline entries (i.e. temporal relations involving $p$), where each timeline entry $r_i$ is relevant to the person biography $bio$.  
\end{definition}

\subsection{\system{} Components}

\system{} consists of several components that together enable interaction with the biography timeline. Fig. \ref{fig:screen} presents an example of the generated biography timeline for John Adams, the second president of the United States.

\textbf{Wikipedia biography.} On top, a brief textual biography and the person's Wikipedia link is shown next to the person's image. 

\textbf{Event map.}
An interactive map displays the locations of timeline entries and events in the person's life.

\textbf{Biography timeline.}
The actual biography timeline is displayed in the centre. At first glance, the user can see the person's life span, as well as relevant phases in the person's life. Among other timeline entries, the example timeline indicates the birth of Adams' child, as well as his term as US president. The user can interact with the timeline to obtain additional information.

\textbf{Related people.}
Below the timeline, a list of people relevant to the selected person is shown to enable the exploration of further biography timelines.

\textbf{Events.}
A chronological list of events in the person's life is presented.

\subsection{User Interaction and Data Export}

The different components of \system{} are connected and are highly interactive. For example, a click on a timeline entry leads to the selection of the associated location, event and people.

\system{} does also offer an export option for the events and relations that underline the timeline generation, which provides access to the timeline facts in a JSON file. Moreover, the exported file contains all the temporal relations that were judged as non-relevant by our model. That way, we envision that \system{} can  facilitate further research on biography timeline generation from the knowledge graph. 

\afterpage{%
\begin{figure}[t]
  \centering
   \fbox{\includegraphics[width=\textwidth]{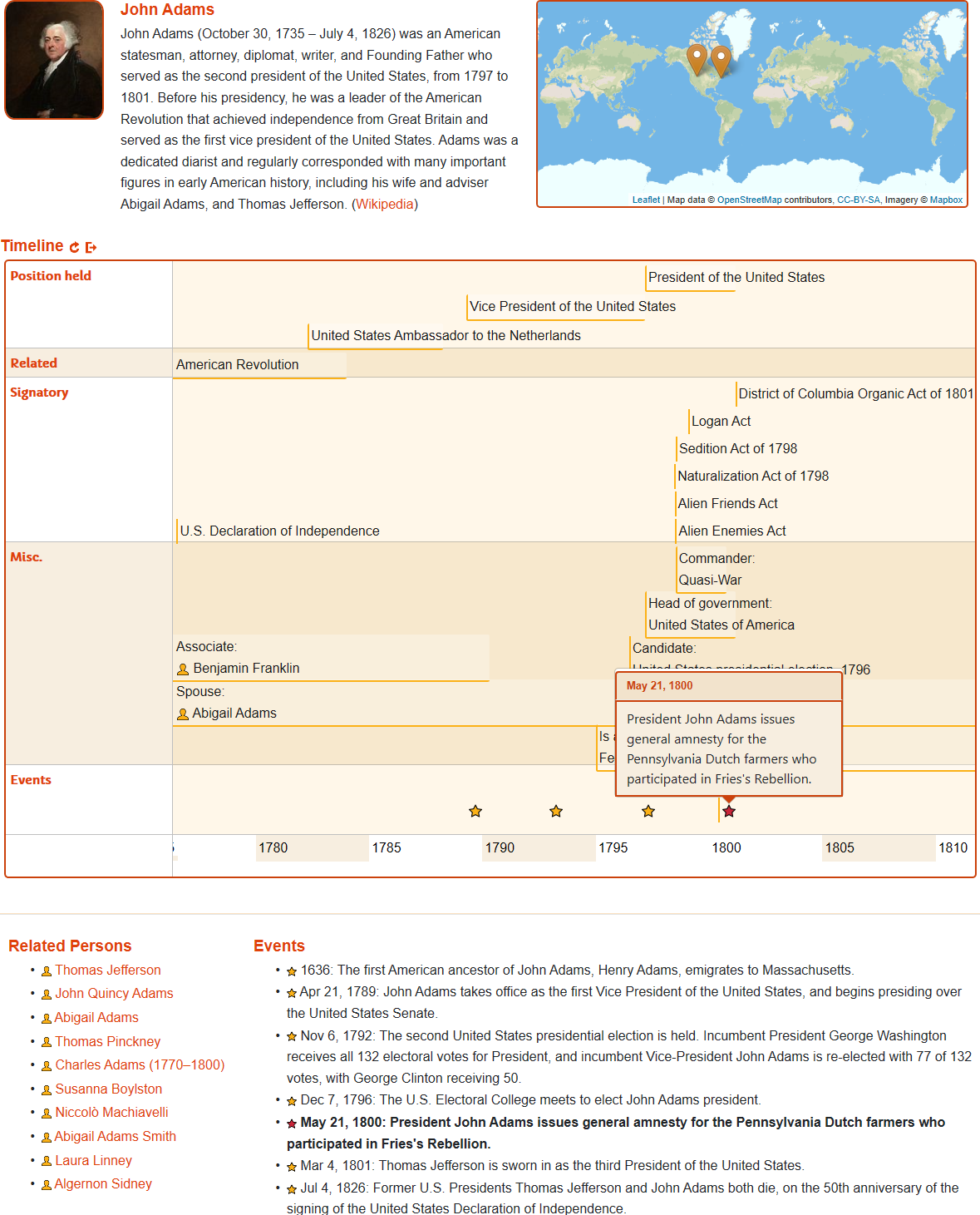}}
  \caption{Biography timeline of John Adams, showing a short textual biography, a map, the generated biography timeline, related people and events. If possible, the timeline entries are grouped by property labels of the underlying temporal relations (e.g., ``Position held'' and ``Signatory'', plus ``Misc.'' for all properties only covered in the timeline once). The ``Events'' section shows textual events related to John Adams, e.g. a sentence about amnesty for farmers.
  }
  \label{fig:screen}
\end{figure}
\clearpage
}
\section{Biography Timeline Generation}
\label{sec:timelines}

The goal of the biography timeline generation approach is to predict whether a temporal relation is relevant to the biography of the person of interest $p$. A temporal relation is a binary relation between $p$ and a connected entity.
This relation includes the property identifier (e.g., ``child'' or ``marriedTo'') and
a validity time interval or time point (e.g., ``March 4, 1797 -- March 4, 1801'').

EventKG does not only provide information about events but also their connections to related entities, as well as temporal relations between entities. In \system{}, we extract temporal relations from EventKG as the union of the following three relation types:

\begin{enumerate}
    \item Relations where the object is a temporal literal. Example:
    \begin{itemize}
        \item \textit{John Adams}, \textit{born}, \textit{30 Oct 1735}
    \end{itemize}
    \item Relations that are directly assigned a validity time span. Example:
    \begin{itemize}
        \item \textit{John Adams}, \textit{marriedTo}, \textit{Abigail Adams} [25 Oct 1764 -- 28 Oct 1818]
    \end{itemize}
    \item Indirect temporal relations where the validity time is identified using the object's happening or existence time. Example:
    \begin{itemize}
       \item \textit{John Adams}, \textit{child}, \textit{John Quincy Adams} [11 Jul 1767 -- 4 Jul 1826]\footnote{John Quincy Adams was born on July 11, 1767 and his father died on July 4, 1826.}
    \end{itemize}
\end{enumerate}

The distant supervision approach to identify relevant temporal relations adopted by \system{} is described in detail in our previous work \cite{gottschalk2019eventkg} and follows three major steps:

\textbf{Benchmark creation.} From two different corpora (Wikipedia abstracts and biographical websites\footnote{\url{https://www.biography.com/}}$^,$\footnote{\url{https://www.thefamouspeople.com/}}), biographies of well-known persons are extracted and mapped to the temporal relations in the EventKG knowledge graph.

\textbf{Feature extraction.} 
From each temporal relation, we extract a set of features characterising the person type, indicating the importance of the connected entity, characterising the relation and its temporal properties.

\textbf{Model training and timeline generation.} Based on a training set of temporal relations marked as (non-)relevant to the benchmark, SVM classifiers are trained to predict the relevance of a temporal relation for a person's biography.
In \system{}, we make use of two SVMs, one trained on the Wikipedia abstracts and another one trained on the biographical websites.

In addition, \system{} also provides textual events (e.g., ``President John Adams issues general amnesty for the Pennsylvania Dutch farmers who participated in Fries's Rebellion'') that are queried from EventKG. 
\section{Datasets and Implementation}
\label{sec:dataset}

\system{} relies on models pre-trained on Wikipedia and biographical websites, temporal relations extracted on-the-fly from EventKG and additional information obtained from Wikipedia (the brief textual biography and image). The user can generate biography timelines for nearly $1.25$ million persons. The pre-trained models were learnt on a benchmark consisting of 2,760 persons and more than $750$ thousand biography entries that is also publicly available\footnote{\url{http://eventkg.l3s.uni-hannover.de/timelines.html}} \cite{gottschalk2019eventkg}.

\system{}\footnote{\url{http://eventkg-biographies.l3s.uni-hannover.de}} is accessible as an HTML5 website implemented using the Java Spark web framework\footnote{\url{http://sparkjava.com/}}. The biography timelines are visualised through the browser-based Javascript library vis.js\footnote{\url{http://visjs.org/timeline_examples.html}}, the maps are generated through the Leaflet Javascript library\footnote{\url{https://leafletjs.com}}, and pop-overs showing detailed information are based on Bootstrap\footnote{\url{https://getbootstrap.com/}}. EventKG data is queried through its SPARQL endpoint\footnote{\url{http://eventkg.l3s.uni-hannover.de/sparql.html}}, and Wikipedia information is retrieved via the MediaWiki action API\footnote{\url{https://www.mediawiki.org/wiki/API:Main_page}}. To reduce the number of calls to the SPARQL endpoint, biography timelines are cached.
\section{Demonstration}
\label{sec:demonstration}

In our demonstration, we will show how \system{} works and how users can use it to generate biography timelines. We will give the users the option to select any person of interest, but also prepare a diverse set of people with interesting timelines (e.g., John Adams, Angelina Jolie, Albert Einstein, Lionel Messi). By comparison with Wikipedia articles, we will demonstrate how \system{} gives a particularly fast first impression of a person's life.
\section{Related Work}
\label{sec:related_work}

\textbf{Timeline Generation and Entity Summarisation.} The biography timelines shown in \system{} are based on our previous work on automated biography timeline generation using distant supervision on relations identified in textual biographies \cite{gottschalk2019eventkg}. In contrast, the biography timelines of Althoff et al. are created based on an optimisation task with hand-crafted constraints and features \cite{althoff:2015}. In a similar setting, traditional entity summarisation (e.g. \cite{diefenbach2018pagerank}) aims at the identification of relevant facts given a query concept. While entity summarisation approaches also utilise semantic information given in knowledge graphs, they are not considering temporal information.

\textbf{Biography and Timeline Visualisation.} Few systems exist that provide visualisations of biography timeline extracted from knowledge graphs: BiographySampo \cite{hyvonen2019biographysampo} provides Finnish textual biographies that a user can explore using network exploration and maps. The TimeMachine by Althoff et al. \cite{althoff:2015} gives a compact overview of only a few related entities but does not provide time intervals, or any further information. EventKG+TL \cite{gottschalk2018eventkgtl} is another system built on top of EventKG that provides event timelines. BiographySampo and TimeMachine are limited to a pre-selected set of person entities. \system{} offers different views for nearly $1.25$ million persons that are accessible through a common interactive interface. Also, \system{} focuses on the provision of relevant information: there is no restriction on a limited amount of relations, but \system{} also does not overwhelm the user with all possible information available.
\section{Conclusion}
\label{sec:conclusion}

In this paper, we discussed how knowledge graphs could facilitate research on notable accomplishments and essential events in the life of people of public interest. We presented \system{} that generates a concise overview of a person's biography on an interactive timeline from the EventKG knowledge graph.

\subsubsection*{Acknowledgements} 
This work was partially funded by the EU Horizon 2020 under MSCA-ITN-2018 ``Cleopatra'' (812997), and the Federal Ministry of Education and Research, Germany (BMBF) under ``Simple-ML'' (01IS18054).


 \bibliographystyle{splncs04}
 \bibliography{bibliography}

\end{document}